\documentclass[conference]{IEEEtran}
\IEEEoverridecommandlockouts
\usepackage{cite}
\usepackage{amsmath,amssymb,amsfonts}

\usepackage{algorithm}
\usepackage{algpseudocode}

\usepackage{graphicx}
\algrenewcommand{\algorithmiccomment}[1]{\hskip1.5em$\triangleright$ #1}
\usepackage{textcomp}
\usepackage{xcolor}
\def\BibTeX{{\rm B\kern-.05em{\sc i\kern-.025em b}\kern-.08em
    T\kern-.1667em\lower.7ex\hbox{E}\kern-.125emX}}
\begin{document}

\title{Beyond Pairwise Feedback: Listwise Vision-Language Supervision for Preference-Based Reward Learning \\

}

% \author{\IEEEauthorblockN{Srivalli Kakuri\textsuperscript{*}}
% \IEEEauthorblockA{\textit{School of Mechanical Engineering} \\
% \textit{Purdue University}\\
% West Lafayette, USA \\
% skatkur@purdue.edu}
% \and

% \IEEEauthorblockN{Maxwell Kawada\textsuperscript{*}}
% \IEEEauthorblockA{\textit{Weldon School of Biomedical Engineering} \\
% \textit{Purdue University}\\
% West Lafayette, USA \\
% mkawada@purdue.edu}
% \and

% \IEEEauthorblockN{Juan Wachs}
% \IEEEauthorblockA{\textit{Edwardson School of Industrial Engineering} \\
% \textit{Purdue University}\\
% West Lafayette, USA \\
% jpwachs@purdue.edu}

\author{
\IEEEauthorblockN{
\makebox[0.32\textwidth][c]{Srivalli Katkuri\textsuperscript{*}}
\makebox[0.32\textwidth][c]{Maxwell Kawada\textsuperscript{*}}
\makebox[0.32\textwidth][c]{Juan Wachs}
}

\IEEEauthorblockA{
\makebox[0.32\textwidth][c]{%
\begin{tabular}[t]{c}
\textit{School of Mechanical Engineering} \\
\textit{Purdue University} \\
West Lafayette, USA \\
skatkur@purdue.edu
\end{tabular}
}
\makebox[0.32\textwidth][c]{%
\begin{tabular}[t]{c}
\textit{Weldon School of} \\
\textit{Biomedical Engineering} \\
\textit{Purdue University} \\
West Lafayette, USA \\
mkawada@purdue.edu
\end{tabular}
}
\makebox[0.32\textwidth][c]{%
\begin{tabular}[t]{c}
\textit{Edwardson School of} \\
\textit{Industrial Engineering} \\
\textit{Purdue University} \\
West Lafayette, USA \\
jpwachs@purdue.edu
\end{tabular}
}
}

\thanks{\textsuperscript{*}Srivalli Katkuri and Maxwell Kawada contributed equally to this work.}
\thanks{This work has been submitted to the IEEE for possible publication.
Copyright may be transferred without notice, after which this version may
no longer be accessible.}

}

% \author{\IEEEauthorblockN{Anonymous}}

\maketitle

\begin{abstract}
Vision-language models (VLMs) have emerged as a powerful source of supervision for reinforcement learning, enabling agents to leverage rich semantic knowledge during training. Inspired by the success of preference-based reward learning (PbRL) in reinforcement learning from human feedback (RLHF), vision-language model generated image-based preferences provide an effective source for learning reward functions. This can be done by visually comparing two outcomes through the Bradley-Terry (BT) model. However, this pairwise formulation utilizes only two observations at a time, despite VLMs being capable of ranking multiple candidates. The Plackett-Luce (PL) formulation can shape a reward model with listwise rankings as opposed to pairwise preferences, allowing for a more suited use of a VLM based ranking. In this work, to our knowledge, we introduce the first framework that combines VLM-generated preferences with the Plackett-Luce model for reward learning. We evaluate our approach on Meta-World manipulation tasks and show that Plackett-Luce (PL) reward models can train robotic policies from VLM-generated rankings as effectively as pairwise Bradley-Terry, $\mathbf{K}$-wise Bradley-Terry, and RL-VLM-F baselines. Across all environments, at least one PL ranking size ($\mathbf{K \in \{3,4,5\}}$) consistently performs with or outperforms other methods in mean success rate. Unlike pairwise methods, which are restricted to $\mathbf{K=2}$, PL supports different ranking sizes and can therefore be adapted to the environment and desired feedback format. Our best PL configuration achieves an 86\% mean final success rate and matches the Oracle baseline on Drawer Open. Moreover, either PL or $\mathbf{K}$-wise Bradley-Terry achieves the highest mean success rate among preference-based methods in two of the three environments. Overall, these results demonstrate that listwise VLM preference supervision is a competitive and flexible approach to reward learning for reinforcement learning.

 %  We evaluate our approach on Meta-World manipulation tasks and demonstrate that Plackett-Luce reward learning can successfully train robotic policies from VLM-generated rankings, comparable to other methods such as pairwise and K-Wise Bradley-Terry and the well known RL-VLM-F. Results show that the Plackett–Luce ranking model is a competitive approach for VLM based PbRL, as it consistently finds . In contrast, the results of the RL-VLM-F and Bradley–Terry Pairwise methods fluctuate across environments, with no flexibility to edit their ranking sizes due to the enforced $\mathbf{K=2}$ restriction. PL allows users to vary the number of samples ranked jointly, providing the flexibility to adapt the feedback format to each environment. Our best Plackett-Luce configuration achieved a mean final success rate of 86\%, matching the Oracle baseline on the Drawer Open task.These results establish the viability of listwise VLM preference supervision for learning reward functions in reinforcement learning.

 % In two of three environments, either listwise Bradley-Terry or Plackett-Luce methods had the highest mean success rate among the preference-based methods.  - i got rid of this sentence
\end{abstract}

\begin{IEEEkeywords}
Reinforcement Learning, Preference-Based Reward Learning, Vision-Language Models, Bradley-Terry, Plackett-Luce, Robotics
\end{IEEEkeywords}

\section{Introduction}
Preference-based reinforcement learning (PbRL) emerged largely as a way to reduce human effort \cite{christiano2017deep}. Writing reward functions by hand is often difficult, as many goals resist formalization, and approximate rewards tend to be optimized in ways the designer never intended \cite{amodei2016concrete, pan2022effects}. PbRL addresses this by asking less of the human. Instead of specifying good behavior, one only needs to recognize it by picking the better of two potential options. A common model for such pairwise preferences is the Bradley-Terry (BT) model \cite{bradley1952rank, hamilton2025routesubiquitousbradleyterrymodel}, and it has been implemented in robotics, healthcare, competitive sports, and more \cite{wang2024rlvlmfreinforcementlearningvision, 11495200, cattelan2013dynamic}. The field has since grown beyond pairs, with reward functions learned directly from a complete listwise ranking using the Plackett-Luce (PL) model \cite{plackett1975analysis, zhu2024principledreinforcementlearninghuman}. 

% Related works have also used the PL model to study subset-wise ranking and active preference elicitation \cite{saha2019activerankingsubsetwisepreferences, thekumparampil2024comparingrankmanyactive, mukherjee2026optimaldesignhumanpreference}.

Nevertheless, labeling still requires human involvement, and gathering it can be both time-consuming and expensive \cite{christiano2017deep, lee2021pebble}. This naturally raises the question of whether humans must be in the loop at all, and whether quality supervision could instead be provided on demand, in real time during training, and at scale. Vision-language models (VLMs) have recently been explored as one such alternative. Demonstrating capability to provide task-conditioned reward signals for reinforcement learning \cite{huang2024vlmrlunifiedvisionlanguage, mahmoudieh2022zeroshot, ma2023liv}, a VLM can inspect and contextually reason about a visual outcome, leveraging the broad knowledge encoded in its LLM backbone. Earlier work used CLIP-style similarity as an online reward signal, but provided limited context beyond image-text or image-image similarity \cite{baumli2024visionlanguagemodelssourcerewards, rocamonde2024vision, sontakke2023roboclip}.

During PbRL, those supplying preferences may struggle to provide a full ranking when differences among multiple candidates are subtle. This is a contributing reason for the popularity of pairwise judgments and the Bradley-Terry model \cite{bradley1952rank} in human-based PbRL, even though theory suggests that listwise Plackett-Luce reward learning is asymptotically more efficient than its pairwise decomposition \cite{zhu2024principledreinforcementlearninghuman} and may offer sample-efficiency gains that grow with ranking size \cite{lee2026preferencebasedreinforcementlearningpairwise}. Existing VLM-preference architectures have continued using this pairwise convention similar to the human-feedback setting \cite{wang2024rlvlmfreinforcementlearningvision, singh2025varpreinforcementlearningvisionlanguage}. However, the cost structure has changed, as VLMs can rank K outcomes in a single query, and they can do so repeatedly without the fatigue that would limit sustained human annotation. At scale, each query covers multiple outcomes rather than a single pair, and this difference may compound across thousands of queries.

To challenge the conventional pairwise methods, we introduce a VLM preference-based reinforcement learning framework that uses VLM-generated listwise rankings to train a Plackett-Luce reward model. Using a series of Meta-World \cite{yu2020meta} robotic environments, we provide the PL model with uniformly sampled observations for GPT-5.6 Luna \cite{openai2026gpt56luna} to provide preference rankings for $K \in 3,4,5$ observations at a time. Comparative experiments are performed with pairwise Bradley-Terry and a variant listwise Bradley-Terry model \cite{lee2026preferencebasedreinforcementlearningpairwise}, along with the popular RL-VLM-F \cite{wang2024rlvlmfreinforcementlearningvision}. Results show that Plackett–Luce ranking is a consistently competitive approach for VLM based PbRL, because in each environment, at least one ranking size ($K \in 3,4,5$) performs in strong contention or the best compared to other baselines. In contrast, the pairwise RL-VLM-F and Bradley–Terry methods have no flexibility to edit their ranking sizes due to the enforced $K=2$ restriction. PL allows users to vary the number of samples ranked jointly, providing the flexibility to adapt the feedback format to each environment.

%Results show that PL is a competitive method for VLM PbRL, as it compares to or outperforms baseline methods in mean success rate.

In summary, we make the following contributions:
\begin{itemize}
\item To our knowledge, we present the first methodology that uses VLM generated listwise rankings to train a Plackett-Luce reward model for reinforcement learning. 
\item We evaluate and validate the Plackett-Luce framework across multiple ranking group sizes against pairwise Bradley-Terry, K-wise Bradley-Terry, and RL-VLM-F.
\item We demonstrate the Plackett-Luce framework's ability to solve a series of robotic rigid object manipulation tasks.
\item We perform an ablation which evaluates the effect of number of feedback groups $M$ per iteration for both Plackett-Luce and K-wise Bradley-Terry.
\end{itemize}

\begin{figure*}[t]
\centering
\includegraphics[width=\textwidth]{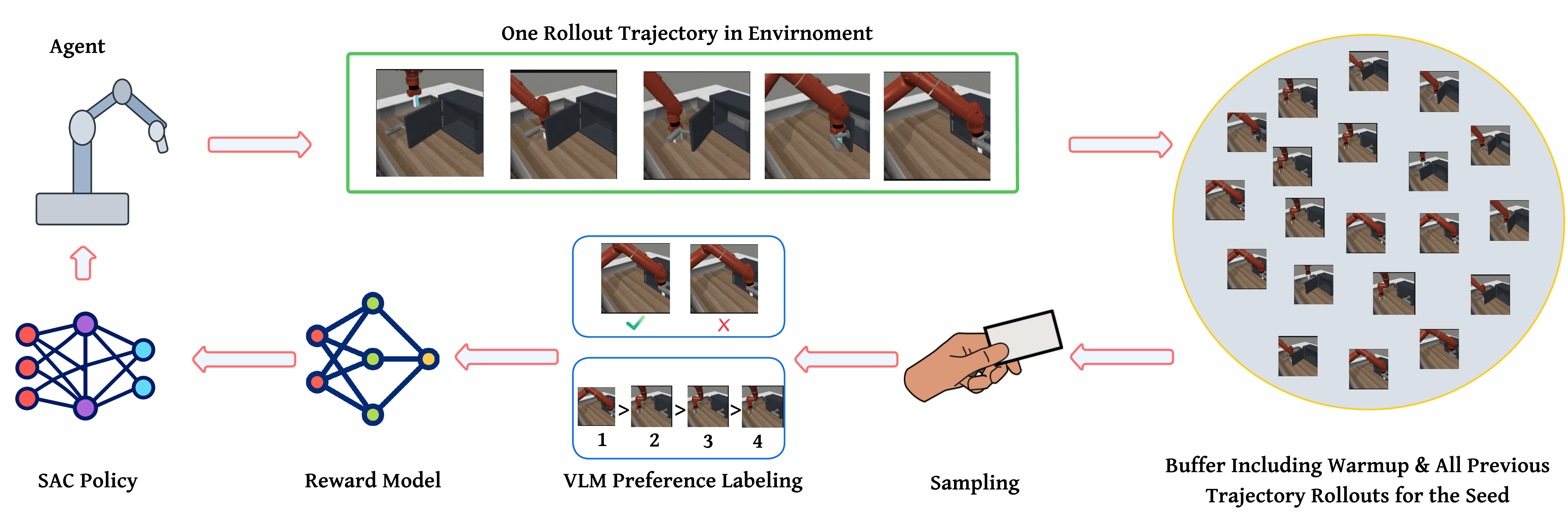}
\caption{System overview for each iteration. Our framework for testing different reward models for VLM preference-based reward learning starts with a simulated robotic arm attempting a task. Scenes from these task attempts are saved into a large buffer, which are then sampled in pairs or groups (pairwise or listwise) for the VLM to generate preference labels. After preference labels are generated, the reward model is updated using either the Plackett-Luce or Bradley-Terry formulation. This reward model is then used to update a Soft Actor-Critic (SAC) policy which dictates how the robotic arm agent acts in the next iteration}
\label{fig:system}
\end{figure*}

\section{Related Work}

\subsection{VLMs for Preference-Based Reward Learning in Robotics}
RL-VLM-F showed that pretrained vision-language models (VLMs) can substitute human annotators by generating pairwise preferences from singular frame observations, conditioned on a natural language task description, which are then used to train a Bradley-Terry reward model. This method outperformed direct VLM raw-score prompting, as well as CLIP- and BLIP-2-based rewards, which compute the reward as the embedding similarity between an image observation and the task description, across all evaluated tasks \cite{wang2024rlvlmfreinforcementlearningvision}. Later work then provided VLMs with richer temporal context to compare, such as visually providing the path taken on final observations \cite{singh2025varpreinforcementlearningvisionlanguage}. However, these comparisons remained pairwise, leaving the richer information from comparing multiple trajectories simultaneously unavailable.  \cite{singh2025varpreinforcementlearningvisionlanguage, wang2024rlvlmfreinforcementlearningvision}.

\subsection{Plackett-Luce Reward Learning}
Zhu et al. \cite{zhu2024principledreinforcementlearninghuman} demonstrated that reward models can be learned directly from Plackett-Luce rankings, and showed theoretically that preserving listwise preference information improves statistical efficiency compared to decomposing rankings into independent pairwise comparisons. Subsequent work focused on extending Plackett-Luce preference learning to active preference elicitation and RLHF-motivated ranking settings, while continuing to rely on human preferences\cite{mukherjee2026optimaldesignhumanpreference,thekumparampil2024comparingrankmanyactive}. HALO is an early example of PL-style reward learning applied to vision-based robotics beyond language models, learning a reward over ranked action sets from egocentric views and outperforming both hand-engineered rewards and vision-based navigation baselines in real-world trials \cite{seneviratne2025halohumanpreferencealigned}. Nevertheless, existing PL results largely assume human-provided preferences. It remains underexplored whether the benefits of PL persist when preferences are generated by a VLM.

\subsection{Comparison of Bradley-Terry and Plackett-Luce}
Lee, Yi, and Oh demonstrate that, under the PL model, ranking feedback can theoretically yield better sample efficiency as the ranking length increases, provided that suitably informative observations are selected for the analysis. Previous listwise analyses did not manage to obtain this improvement. At the same time, they also show that a rank-breaking method based on the Bradley-Terry model can deliver better computational efficiency and empirical performance compared to the PL method when using the same listwise feedback \cite{lee2026preferencebasedreinforcementlearningpairwise}. On the other hand, Xu and Kankanhalli find that both Bradley-Terry and Plackett-Luce become fragile under nearly deterministic preferences, where small perturbations can cause disproportionate changes in other comparisons. However, they prove the listwise PL model is strictly more robust to this sensitivity than Bradley-Terry \cite{xu2025strong}. Taken together, these results still indicate certain theoretical advantages of PL.

\section{Background}

We consider the standard Markov decision process and reinforcement learning setup \cite{sutton1998reinforcement}. At every timestep $t$, the agent receives a state $s_t$ from the environment and chooses an action $a_t$ according to a policy $\pi(a_t \mid s_t)$. The environment returns a reward $r_t$ and transitions to $s_{t+1}$. The goal of the agent is to maximize the return, the discounted sum of rewards $R = \sum_{k=0}^{\infty} \gamma^k r(s_k, a_k)$ with discount factor $\gamma$.

\subsection{Preference-based reinforcement learning}
In preference-based reinforcement learning (PbRL), the reward function is not given. Instead, it is learned from preference labels over the agent's own behaviors \cite{lee2021pebble, christiano2017deep}. A segment $\sigma$ is a sequence of states and actions produced by the agent, stored in a buffer $\mathcal{B}$ as training proceeds. Since a buffer of $N$ segments means $\binom{N}{2}$ possible pairs, and an even greater amount of possible groups, labeling every comparison is infeasible. So, PbRL methods instead budget $M$ comparisons per preference iteration \cite{lee2021pebble, wang2024rlvlmfreinforcementlearningvision}. Given a pair of segments $(\sigma^0, \sigma^1)$, the labeler indicates which segment is preferred, recorded as a label $y \in \{0, 1\}$ where $y = 1$ denotes $\sigma^1 \succ \sigma^0$. Labelers also may have the option to assign $y=1$, indicating that there is no discernible preference among the pair of segments \cite{wang2024rlvlmfreinforcementlearningvision}. The Bradley-Terry (BT) model \cite{bradley1952rank, christiano2017deep} fits preferences to a parameterized reward function $\hat{r}_\psi$:

\begin{equation}
P_\psi[\sigma^1 \succ \sigma^0] =
\frac{\exp \sum_t \hat{r}_\psi(s_t^1, a_t^1)}
     {\sum_{i \in \{0,1\}} \exp \sum_t \hat{r}_\psi(s_t^i, a_t^i)},
\end{equation}
and $\hat{r}_\psi$ is trained by minimizing the negative log-likelihood of the labels in the preference dataset:
\begin{multline}
\mathcal{L}_{\text{BT}} =
- \mathbb{E}_{(\sigma^0, \sigma^1, y) \sim \mathcal{D}_{\text{pref}}}
\Big[ \mathbb{I}\{y = 1\} \log P_\psi[\sigma^1 \succ \sigma^0] \\
+ \mathbb{I}\{y = 0\} \log P_\psi[\sigma^0 \succ \sigma^1] \Big].
\end{multline}
The reward function and policy are updated alternately. $\hat{r}_\psi$ is fit to the growing preference dataset, and the policy is optimized against $\hat{r}_\psi$ using a standard RL algorithm.

\subsection{Plackett-Luce ranking model}
The Plackett-Luce (PL) model \cite{plackett1975analysis, zhu2024principledreinforcementlearninghuman} generalizes Bradley-Terry from pairs to rankings. Given $K$ segments ranked $\sigma^{(1)} \succ \dots \succ \sigma^{(K)}$, the ranking is modeled as repeatedly choosing the best remaining segment, giving the likelihood
\begin{equation}
P_\psi\big[\sigma^{(1)} \succ \dots \succ \sigma^{(K)}\big] =
\prod_{k=1}^{K}
\frac{\exp \hat{r}_\psi(\sigma^{(k)})}
     {\sum_{j=k}^{K} \exp \hat{r}_\psi(\sigma^{(j)})},
\end{equation}
where $\hat{r}_\psi(\sigma) = \sum_t \hat{r}_\psi(s_t, a_t)$, and training again minimizes the negative log-likelihood. For $K = 2$, PL reduces exactly to BT. A $K$-wise ranking can also be rank-broken, decomposed into its $\binom{K}{2}$ implied pairwise comparisons and fit with the BT loss, which keeps the data but discards the joint structure of the ranking \cite{zhu2024principledreinforcementlearninghuman}.

\section{Methods}

\subsection{Data collection}

Our experiments follow the standard preference-based RL loop. As an agent interacts with the environment, which in our case is simulated, a reward model is trained from preferences over its collected observations, and the policy is optimized against the rewards that this model assigns. We use SAC \cite{haarnoja2018soft} as the policy-learning algorithm throughout. As an off-policy method, it trains on transitions drawn from a replay buffer rather than only on freshly collected experience. We adopt SAC because it is widely used for continuous-control robotic manipulation, making our results directly comparable to prior preference-based RL work in this setting \cite{lee2021pebble, wang2024rlvlmfreinforcementlearningvision}.

At each training iteration, we reset the environment and collect a single trajectory from the resulting initial configuration, with each trajectory containing at most 500 environment steps. We do not exploit the simulator's ability to restore an identical state and generate multiple counterfactual trajectories from the same starting point. Each interaction round therefore contributes exactly one newly executed trajectory.

This choice is motivated by the longer-term goal of making preference-based reward learning more applicable to physical robot learning. In simulation, restoring the same state and collecting several alternative trajectories is relatively inexpensive. In a physical environment, however, reproducing an identical starting state can be difficult, and repeatedly executing exploratory trajectories can require substantial time and physical intervention. Such interactions may also be dangerous early in training when the policy is unreliable. We therefore investigate whether a reward model and policy can improve while collecting only one new trajectory during each interaction round.

Before VLM preference learning begins, we perform iterations where SAC produces rollouts with a state-entropy exploration objective rather than rewards \cite{lee2021pebble}. The warm up provides an initial pool of behavior for the first preference query. Otherwise, the first query to the VLM would be restricted to observations from approximately one trajectory and would contain little behavioral coverage.

We recognize that this initialization is in tension with our real-world motivation because during warm up, the policy may execute undesirable exploratory behavior because feedback is not yet received for training. We retain the warm up in the present experiments because they are conducted in simulation, and it allows us to study the one-trajectory-per-iteration protocol without simultaneously introducing the additional cold-start problem of learning from a nearly empty buffer. 
% Determining whether the warm up can be shortened or removed, so that preference learning begins from the first collected trajectory, is a direction for future work.

Every transition and its associated rendered observation are stored in a replay buffer with a fixed capacity of frames. Fixing the capacity prevents memory, storage, and reward-processing costs from growing without bound while retaining a large pool of experience for SAC training. For preference acquisition, the buffer is treated as a mixed pool rather than being organized by trajectory identity or timestep. Observations are sampled uniformly, meaning that every stored observation has an equal chance of being selected, without using a predetermined strategy to identify especially informative examples. Therefore, observations appearing in the same query for evaluation by the VLM for the reward model may originate from the same trajectory or from different trajectories. Additionally, when the VLM compares observations from different trajectories for the reward model, those observations do not need to correspond to the same timestep or the same level of task progress, and observations drawn from a single trajectory can be separated by any length of time. The buffer grows by one trajectory per training iteration. The preference-learning methods begin with the experience accumulated during warm-up and progressively gain access to a broader distribution of policy behavior, and this steadily growing dataset is continually reused by SAC's off-policy updates.

\begin{algorithm}[t]
\caption{Unsupervised warm-up}
\label{alg:warm-up}
\raggedright
\begin{algorithmic}[1]
\Require environment $\mathcal{E}$, policy $\pi_\phi$, empty replay buffer $\mathcal{B}$, warm-up iterations $N_{\text{warm}}$, horizon $T$
\For{$i = 1$ to $N_{\text{warm}}$}
    \State reset $\mathcal{E}$ and collect one trajectory $\tau = \{(s_t, a_t, s_{t+1}, o_t)\}_{t=1}^{T}$ with $\pi_\phi$, where $o_t$ is the rendered frame at step $t$
    \State $\mathcal{B} \gets \mathcal{B} \cup \tau$
    \State compute intrinsic state-entropy rewards $r^{\text{ent}}_t$ from $k$-nearest-neighbor distances between states in a minibatch sampled from $\mathcal{B}$
    \State update $\pi_\phi$ with SAC using $r^{\text{ent}}$, with no reward model queried or trained
\EndFor
\State \Return $\mathcal{B}$, $\pi_\phi$
\end{algorithmic}
\end{algorithm}

\begin{figure*}[t]
\centering
\includegraphics[width=\textwidth]{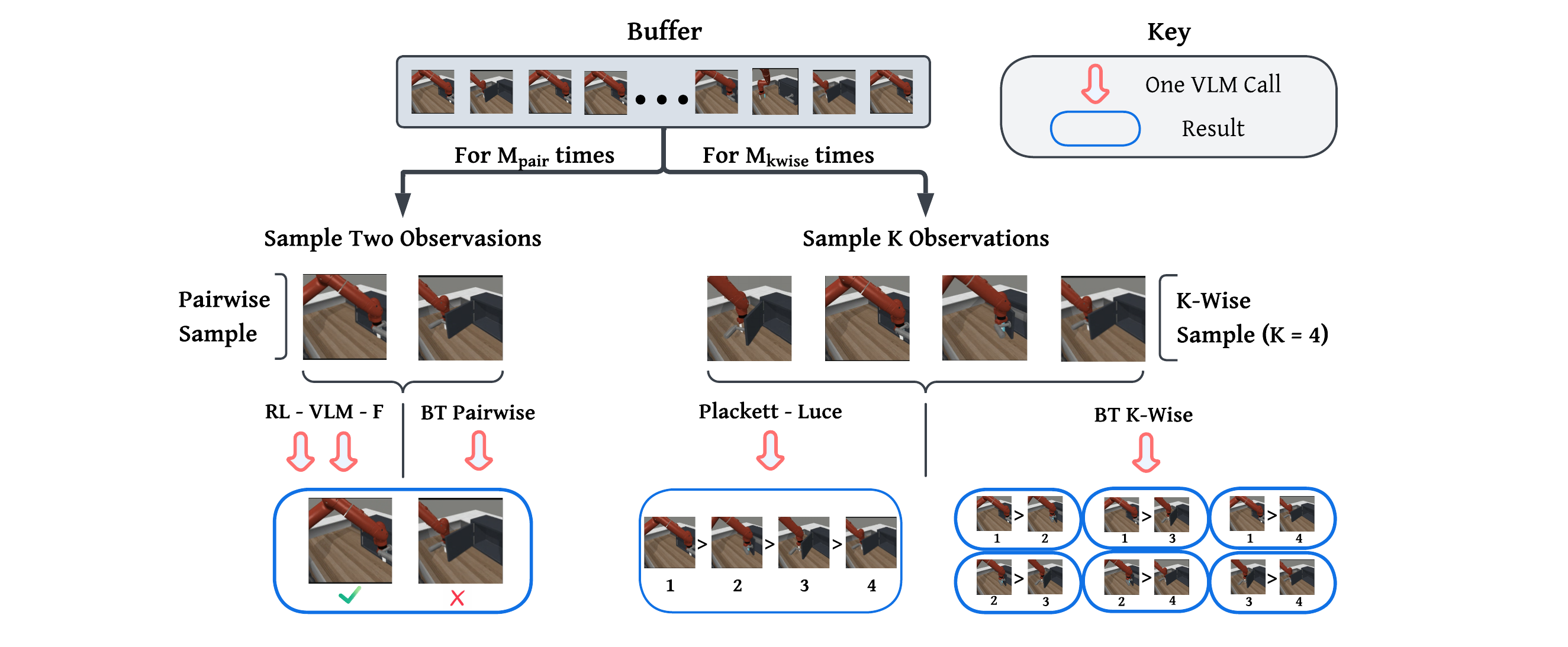}
\caption{Visual description of each sampling method to obtain preference labels. From the buffer of images collected throughout iterations, either two observations or K observations are sampled at the same time. K refers to the number of observations that the VLM is simultaneously queried with for listwise methods. Pairwise samples are evaluated by the VLM using either the RL-VLM-F or BT-Pairwise method. The RL-VLM-F method uses two VLM queries, whereas BT-Pairwise does the same comparison with one VLM query. Listwise samples are evaluated by the VLM using either the Plackett-Luce or BT-Kwise method. Both of these methods require a single VLM query to obtain preference labels.}
\label{fig:sampling}
\end{figure*}

\subsection{Listwise comparison protocol}
Our primary research question is whether an effective visual reward model can be trained from listwise VLM preferences using the Plackett-Luce objective. We evaluate $K \in \{3,4,5\}$, where each active preference-learning iteration issues $M$ independently sampled listwise ranking queries. For every query, $K$ distinct observations are uniformly sampled from the replay buffer, presented together to the VLM, and ranked according to their visual task progress. Each stored observation in the buffer has an equal chance of being selected, and no acquisition strategy is used to identify especially informative groups. The resulting complete rankings are then used to train the Plackett-Luce reward model.

We vary $K$ because increasing the ranking size introduces a potential tradeoff. A longer ranking communicates relationships among more observations and contains more implied comparisons. At the same time, a request containing more observations may include visually similar states whose relative progress is difficult to distinguish, because the VLM must reason jointly over a larger visual context. We therefore do not presume that a larger $K$ is necessarily better and instead evaluate which ranking sizes produce the strongest learned rewards and downstream policies.

For every value of $K$, we evaluate a corresponding BT-Kwise condition. PL and BT-Kwise use the same ranking size, number of VLM calls per iteration, uniform sampling procedure, and complete-ranking VLM requests. However, following the ranking, PL trains from its joint likelihood, whereas BT-Kwise decomposes it into all $\binom{K}{2}$ implied pairwise relations and trains using the BT objective. Rankings of three, four, and five observations therefore produce three, six, and ten implied pairwise relations, respectively.

Building on this, we also report preliminary findings where PL operates with $M=1$, issuing a single $K$-wise ranking query per active iteration, evaluated against the same pairwise baselines at their full budgets. Under this protocol, PL uses one VLM request per iteration, BT-Pairwise uses four, and RL-VLM-F uses eight. This comparison examines whether one structured ranking can substitute for multiple pairwise queries. 

% We perform five independent seeds for every configuration of PL and BT-Kwise. Each matched pair of conditions shares the same ranking size $K$ and the same number of ranking queries $M$ issued per iteration, which controls both the number of observations presented per request and the number of ranking requests issued. Consequently, this design reduces confounding differences between PL and BT-Kwise and isolates the reward-learning objective as the primary methodological distinction.
\begin{algorithm}[t]
\caption{\textsc{KwiseFeedback}}
\label{alg:kwise}
\raggedright
\begin{algorithmic}[1]
\Require buffer $\mathcal{B}$, preference dataset $\mathcal{D}$, queries per iteration $M$, ranking size $K \in \{3,4,5\}$, method $m$
\For{each of the $M$ queries}
    \State draw $K$ distinct frames $\{o^1, \dots, o^K\}$ uniformly from $\mathcal{B}$, independently across queries
    \State complete ranking: $\sigma \gets \textsc{VLM}_{\text{rank}}(o^1, \dots, o^K)$
    \If{$\sigma$ is not a valid ranking of the $K$ frames}
        \State discard the query
    \ElsIf{$m = \text{PL}$}
        \State store the full ranking: $\mathcal{D} \gets \mathcal{D} \cup \{(\{o^1,\dots,o^K\}, \sigma)\}$
    \ElsIf{$m = \text{BT-Kwise}$}
        \State decompose $\sigma$ into all $\binom{K}{2}$ implied pairwise relations
        \State $\mathcal{D} \gets \mathcal{D} \cup \{(o^p, o^q, y^{pq}) : p < q\}$
    \EndIf
\EndFor
\State \Return $\mathcal{D}$
\Statex \textbf{Cost:} $M$ VLM calls for both methods
\Statex \textbf{Objective:} PL fits $\hat{r}_\psi$ with the Plackett--Luce likelihood over full rankings, BT-Kwise fits $\hat{r}_\psi$ with the Bradley--Terry loss over the decomposed pairs
\end{algorithmic}
\end{algorithm}
\subsection{Pairwise baselines}
We additionally compare PL against two pairwise-feedback baselines. The first is RL-VLM-F \cite{wang2024rlvlmfreinforcementlearningvision}, the original framework for pairwise preference labeling with VLMs. For each pair of candidates, it issues two VLM calls: one to produce a free-form analysis and another to assign the final preference label. The second baseline is a BT-Pairwise configuration, modeled after the standard pairwise-feedback protocol in preference-based reinforcement learning. This method uses a single VLM query to evaluate one pair, directly choosing the better of the two options in a manner analogous to a human or automated teacher.

\begin{algorithm}[t]
\caption{\textsc{PairwiseFeedback}}
\label{alg:pairwise}
\raggedright
\begin{algorithmic}[1]
\Require buffer $\mathcal{B}$, preference dataset $\mathcal{D}$, queries per iteration $M$, method $m$
\For{each of the $M$ queries}
    \State draw two distinct frames $(o^1, o^2)$ uniformly at random from $\mathcal{B}$, independently across queries
    \If{$m = \text{BT-Pairwise}$}
        \State $y \gets \textsc{VLM}_{\text{label}}(o^1, o^2)$
    \ElsIf{$m = \text{RL-VLM-F}$}
        \State free-form analysis: $a \gets \textsc{VLM}_{\text{analyze}}(o^1, o^2)$
        \State preference label: $y \gets \textsc{VLM}_{\text{label}}(o^1, o^2, a)$
    \Else
        \State $y \gets \text{null}$
    \EndIf
    \If{$y$ indicates a preference for $o^1$ or $o^2$}
        \State $\mathcal{D} \gets \mathcal{D} \cup \{(o^1, o^2, y)\}$
    \EndIf
\EndFor
\State \Return $\mathcal{D}$
\Statex \textbf{Cost:} $M$ VLM calls for BT-Pairwise, $2M$ for RL-VLM-F
\Statex \textbf{Objective:} both methods fit $\hat{r}_\psi$ with the Bradley--Terry loss over $\mathcal{D}$
\end{algorithmic}
\end{algorithm}

% Our main experiment compares these baselines against PL while holding the number of sampling events per iteration fixed. In this configuration, PL issues $M=4$ $K$-wise ranking queries per active iteration, each drawing $K$ observations from the replay buffer, mirroring the $M=4$ pairwise queries issued by BT-Pairwise. This design serves two purposes. First, it examines whether a reward model can be trained from VLM-generated $K$-wise rankings at all, independent of any query-efficiency argument. Second, it isolates the effect of feedback structure, because the number of sampling events per iteration is held equal. This allows performance difference to be attributable to what each query communicates rather than to how many queries are issued. A $K$-wise request asks the VLM to jointly order $K$ observations and therefore conveys a full ordering, implying $\binom{K}{2}$ pairwise relations per response, compared to one per pairwise query. If matched-budget PL achieves comparable or stronger downstream performance, it would suggest that list-wise ranking feedback can extract more supervision per query than independent pairwise comparisons. More excitingly, and likely more relevant to the community given RL-VLM-F’s practical use in robotics, this matched-budget setup may offer a cost benefit over RL-VLM-F. Its two-stage prompting scheme requires eight VLM calls to label the same four comparisons. If PL achieves comparable or better downstream policies with only half as many labeling calls, that would be a clear advantage.

\subsection{Uniform query sampling}
We use uniform replay-buffer sampling for all preference-learning configurations: PL, BT-Kwise, BT-Pairwise, and RL-VLM-F. This ensures that no method receives an advantage from uncertainty, disagreement, or diversity-based acquisition, while allowing us to test whether listwise ranking feedback remains effective without a query-optimization stage. This protocol departs from common BT-based practice, in which reward learning often benefits from obtaining informative pairs rather than arbitrary ones \cite{lee2021pebble, metcalf2024sample}.

\begin{algorithm}[t]
\caption{Preference-Based Reward Learning Loop}
\label{alg:main}
\raggedright
\begin{algorithmic}[1]
\Require $\mathcal{B}$, $\pi_\phi$ from Alg.~\ref{alg:warm-up}, reward model $\hat{r}_\psi$, empty preference dataset $\mathcal{D}$
\Require feedback method $m \in \{$RL-VLM-F, BT-Pairwise, BT-Kwise, PL$\}$
\Require queries per iteration $M$, group size $K$, iterations $N$
\For{$i = 1$ to $N$}
    \State reset $\mathcal{E}$ and collect \textbf{one} trajectory $\tau$ with $\pi_\phi$
    \State $\mathcal{B} \gets \mathcal{B} \cup \tau$
    \If{$m \in \{$RL-VLM-F, BT-Pairwise$\}$}
        \State $\mathcal{D} \gets
        \textsc{PairwiseFeedback}(\mathcal{B}, \mathcal{D}, M, m)$
    \Else
        \State $\mathcal{D} \gets
        \textsc{KwiseFeedback}(\mathcal{B}, \mathcal{D}, M, K, m)$
    \EndIf
    \State update $\hat{r}_\psi$ on all accumulated labels in $\mathcal{D}$
    \State relabel rewards of transitions in $\mathcal{B}$ with $\hat{r}_\psi$
    \State update $\pi_\phi$ with SAC using $\hat{r}_\psi$
\EndFor
\State \Return $\pi_\phi$, $\hat{r}_\psi$
\end{algorithmic}
\end{algorithm}

\section{Experiments}
\subsection{Environments and Tasks}
% MetaWorld, selected tasks, horizon, success definition,
% observations, cameras, and task-specific rendering.

To test our framework, we utilized the Meta-World simulation environment to benchmark each method's ability to train a robotic action policy with VLM PbRL \cite{yu2020meta}. Within Meta-World, we are able to train a Sawyer robotic arm to perform various object manipulation tasks. In particular, there are 50 tasks available that span a range of object interaction and motor control objectives. Furthermore, Meta-World is a widely used benchmark in both traditional and preference-based reinforcement learning for single- and multi-task policy training\cite{wang2024rlvlmfreinforcementlearningvision, mclean2026meta}.

Because rewards do not originate from the simulated environment, but rather come from visual abstraction, we must select tasks where completion can be interpreted from the scene itself. Furthermore, task completion should ideally be easily interpreted by an external viewer, making tasks which have minute visual changes after completion undesirable. With this in mind, we selected the following three tasks:
\begin{enumerate}
    \item \emph{Drawer Open}: The robot end-effector must insert into the handle and pull the drawer open
    \item \emph{Door Close}: The robot must push the initially open door shut
    \item \emph{Button Press}: The robot must move horizontally to press a button
\end{enumerate}

\begin{figure}[H]
\centering
\includegraphics[width=\columnwidth]{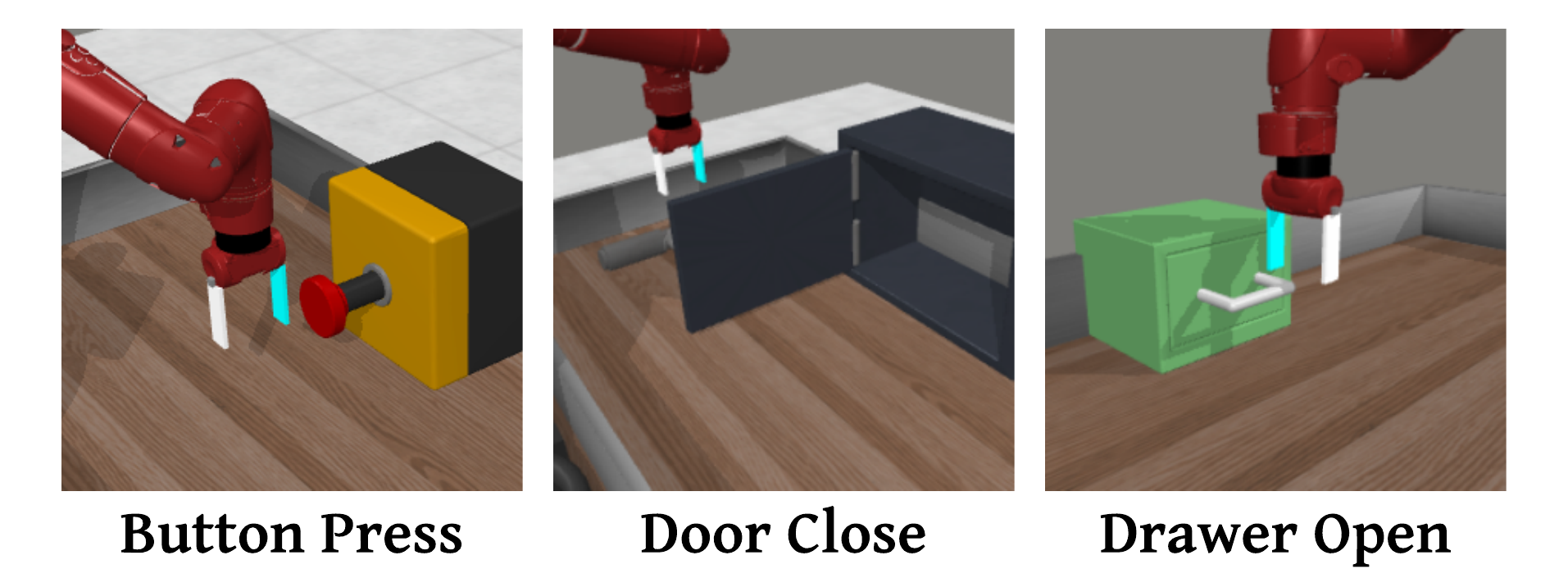}
\caption{Initial position of Sawyer robot arm for each Meta-World task.}
\label{fig:initial}
\end{figure}

These tasks were selected to observe a range of diverse movements and completion markers. This prevents us from testing the same skills across different, but similar, environments. For each of the tasks we use a custom camera position that provides a more object-focused view which can be seen in \ref{fig:initial}. To keep the VLM focused on task completion and to avoid the robot obstructing the view, the robot model is transparent during training. 
% Furthermore, the selection of 'drawer open' as a task allows us to directly compare to 

\subsection{Primary Setup}
% M=4, K={3,4,5}, seeds, 100k steps, uniform buffer sampling,
% VLM model/prompt, reward CNN, SAC, and warm-up.
To evaluate each method with equal number of independently sampled feedback groups, we limit the number of feedback groups to $M=4$ following the warm-up phase. For PL and BT-Kwise, each feedback group is provided $K$ images and requires only one VLM request to return a full ranking. On the other hand, the pairwise methods are fundamentally limited to $K=2$ images, with BT-Pairwise using one VLM request per image pair and RL-VLM-F using $2M=8$ VLM requests due to its two-stage analysis and labeling procedure. We also evaluate the effect of ranking size, $K$, by evaluating both PL and BT-Kwise with $K\in\{3,4,5\}$. Thus, every task and method configuration is evaluated using five independent seeds numbered 0-4.

Each run consists of 100,000 environment steps which is divided into 200 training iterations of 500 steps. The first 18 iterations are used for warm-up, using a PEBBLE-style entropy exploration reward \cite{lee2021pebble}. During warm-up, no VLM queries or ground truth rewards are used and the SAC critics are reset following the warm-up. The next 182 iterations sample images uniformly from a persistent replay buffer with a capacity of 100,000 images. 

We use GPT-5.6 Luna as our VLM, configured with original image detail, and reasoning effort set to none \cite{openai2026gpt56luna}. GPT-5.6 Luna was selected for its balance of high performance and cost-effectiveness. For each task, a goal-specific description is provided to the VLM which is consistent across all methods. These descriptions direct the VLM to give preferences to images which are more indicative of task progress, such as drawer displacement, door angle, and button depression. Prompt and hyperparameter details will be provided in the appendix.

\subsection{Ablation Setup}
% Drawer Open, M=1 versus M=4 at fixed K=3.
% State that all other hyperparameters are held constant.
We also investigated how changing the number of feedback groups from $M=4$ to $M=1$ impacts the performance of the listwise methods, with all other hyperparameters kept consistent. With a smaller number of feedback groups, each run would be querying the VLM a quarter of the time, showcasing a more cost-effective alternative to the main setup. This experiment is only performed on the 'drawer open' task with the number of sampled images fixed at $K=3$. 

\subsection{Evaluation Protocol}
% Evaluation every 10k steps, 20 episodes, deterministic policy,
% final evaluation, success-rate definition, AUC, mean/SEM/CI,
% and seed as the statistical unit.
Policies are evaluated every 10,000 environment steps, including at the final 100,000 step checkpoint, using 20 deterministic Meta-World episodes. Task success is measured using Meta-World's ground-truth success criterion, and the success rate is defined as the number of successful episodes out of 20.

Because all methods were evaluated using the same five matched random seeds, comparisons between PL and BT-Kwise at the same ranking size K were performed using two-sided paired t-tests, with seed as the statistical unit. A Holm correction was applied following the comparisons.

\subsection{Implementation and Reproducibility}
% Hardware, software versions, API configuration, query budgets,
% seeds, W&B logging, and repository commit.
To run experiments, we used the following configuration:

\begin{table}[H]
\centering
\caption{Hardware and Software Versions}
\resizebox{0.9\columnwidth}{!}{%
\begin{tabular}{|l|l|}
\hline
\textbf{OS} & Ubuntu 24.04.3 LTS \\ \hline
\textbf{GPU} & NVIDIA GeForce RTX 3090 (24 GB) \\ \hline
\textbf{NVIDIA Driver} & 580.95.05 \\ \hline
\textbf{NVCC Version} & V12.0.140 \\ \hline
\textbf{CPU} & AMD Ryzen 9 7950X \\ \hline
\textbf{Python} & 3.12.13 \\ \hline 
\end{tabular}%
}
\label{tab:hardware}
\end{table}

Alternative run configurations will be provided in the appendix. All code related to this project will be uploaded to a publicly accessible GitHub repository.

\begin{figure*}[t]
\centering
\includegraphics[width=\textwidth]{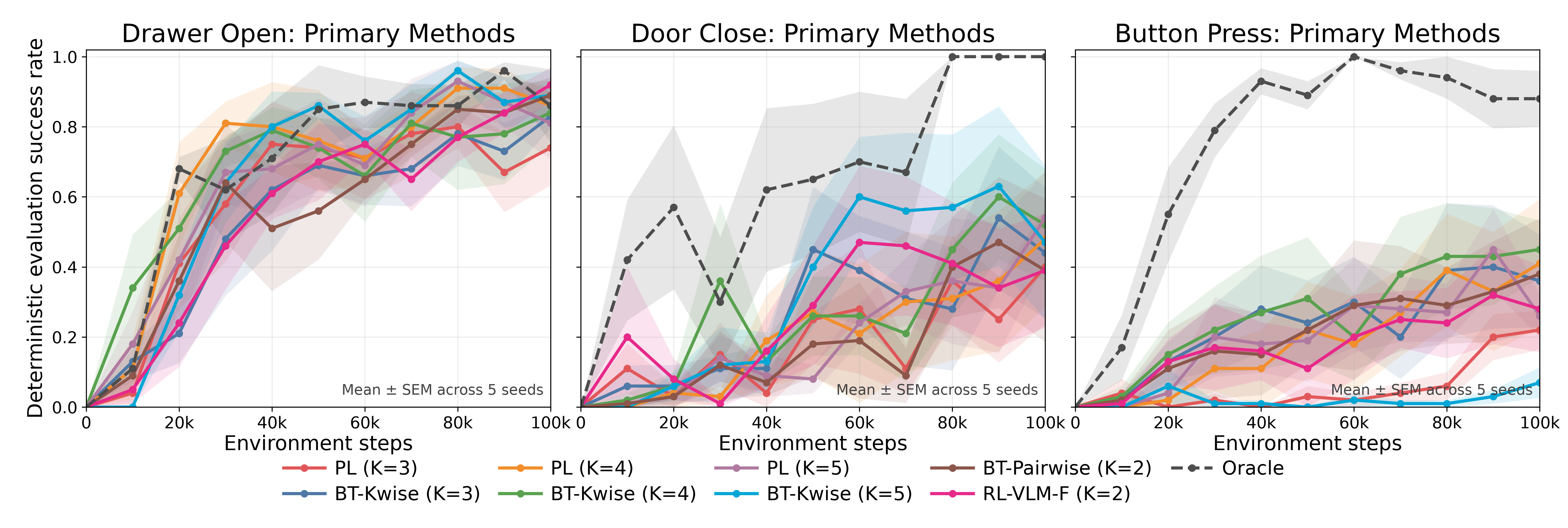}
\caption{Deterministic evaluation success rates for all primary methods and oracle baseline. For the Drawer Open task, RL-VLM-F was able to score the highest final success rate at 92\%, with BT-Pairwise and BT-Kwise ($K=5$) tied for 2nd best at 89\%. Both of these methods outperformed the oracle baseline, which tied with PL ($K=4$) at 86\%. For the Door Close task, PL ($K=5$) was the best performing method, followed by BT-Kwise ($K=4$) at 52\%. However, the oracle method for door close outperformed all methods with 100\% success rate. In the Button Press task, BT-Kwise $K=4$ performed the best with a mean 45\% success rate, followed by PL $K=4$ with a mean of 41\%. Once again, the oracle baseline outperformed all methods in this environment.}
\label{fig:Success}
\end{figure*}

\section{Results}
\subsection{Primary Method Comparison at $M=4$}
% Compare PL, BT-Kwise, BT-Pairwise, and RL-VLM-F.
% Discuss learning curves and final success across tasks.

\begin{table}[H]
\centering
\caption{Primary Success Rates (\%, Mean $\pm$ SEM across five seeds)}
\label{tab:primary_success}
\resizebox{\columnwidth}{!}{
\begin{tabular}{lccc}
\hline
\textbf{Method} &
\textbf{Drawer Open} &
\textbf{Door Close} &
\textbf{Button Press} \\ \hline

PL ($K=3$)
& $74 \pm 10.7$
& $40 \pm 16.4$
& $22 \pm 5.6$ \\

PL ($K=4$)
& $86 \pm 3.7$
& $48 \pm 19.3$
& $41 \pm 18.4$ \\

PL ($K=5$)
& $81 \pm 10.2$
& $\mathbf{54 \pm 13.2}$
& $26 \pm 6.2$ \\ \hline

BT-Kwise ($K=3$)
& $83 \pm 3.4$
& $44 \pm 18.7$
& $36 \pm 13.2$ \\

BT-Kwise ($K=4$)
& $84 \pm 9.1$
& $52 \pm 16.0$
& $\mathbf{45 \pm 8.2}$ \\

BT-Kwise ($K=5$)
& $89 \pm 7.1$
& $47 \pm 22.0$
& $7 \pm 4.4$ \\ \hline

BT-Pairwise
& $89 \pm 5.1$
& $39 \pm 20.5$
& $38 \pm 15.9$ \\

RL-VLM-F
& $\mathbf{92 \pm 4.9}$
& $39 \pm 16.2$
& $28 \pm 12.3$ \\ \hline

Oracle
& $86 \pm 10.4$
& $100 \pm 0.0$
& $88 \pm 8.0$ \\ \hline

\end{tabular}
}
\end{table}

% Plackett-Luce methods were able to consistently perform with, and at times, outperform every other framework as seen in Figure \ref{fig:Success}. Table \ref{tab:primary_success} shows that during the drawer open task, PL with $K=4$ was able to match the performance of Meta-World's oracle training protocol which utilizes ground truth information from the task environment, averaging an 86\% success rate. Though, RL-VLM-F and BT-Pairwise had the best success among the methods with success rates of 92\% and 89\% respectively. Moreover, PL with $K=5$ was the highest performing preference based reward method for the door close task. Interestingly, both K-wise methods outperformed the pairwise baselines, however no method could compare to the oracle performance. 

% To isolate the effect of the reward-learning objective, we compared PL and BT-Kwise at each matched ranking size using two-sided paired t-tests across the five corresponding seeds. No statistically significant difference was observed at any ranking size for Drawer Open ($K=3:p=0.436$, $K=4:p=0.883$, $K=5:p=0.607$) or Door Close ($K=3:p=0.836$, $K=4:p=0.839$, $K=5:p=0.830$). Thus, within the present five-seed evaluation, preserving the complete PL ranking likelihood did not yield a detectable final-success advantage over rank-breaking the same K-wise feedback into BT comparisons. 

A full success rate table for each seed and method is provided in the Appendix.

\subsubsection{Drawer Open}
For the Drawer Open environment under uniform sampling with $M=4$, the Plackett-Luce estimator performed best with $K=4$, achieving an average success rate of $86\%$, compared with $74\%$ for $K=3$ and $81\%$ for $K=5$. The $K=4$ configuration was also the most consistent across seeds, with success rates of $85\%$, $85\%$, $80\%$, $80\%$, and $100\%$. At the same ranking size of $K=4$, PL slightly outperformed the BT-Kwise estimator, which achieved $84\%$ success. However, RL-VLM-F achieved a higher mean success rate of $92\%$ under its corresponding $M=4$ setting, whereas the standard single-stage BT-Pairwise baseline achieved $89\%$. 

\subsubsection{Door Close}
Among the PL configurations evaluated with uniform sampling and $M=4$, $K=5$ achieved the highest mean success rate, reaching $54\%$, compared with $48\%$ for $K=4$ and $40\%$ for $K=3$. PL with $K=5$ also outperformed RL-VLM-F under the corresponding $M=4$ feedback budget, achieving $54\%$ versus $39\%$ mean success. For reference, the standard single-stage BT-Pairwise baseline also achieved $39\%$ success. At the same ranking size of $K=5$, PL exceeded BT-Kwise by $7$ percentage points ($54\%$ versus $47\%$). Overall, the results indicate that jointly ranking five candidates provided the strongest average performance and stability for PL in the door-close environment, while also substantially outperforming the RL-VLM-F baseline under the tested $M=4$ setting. 

\subsubsection{Button Press}
Across the five Button Press seeds with $M = 4$ and uniform sampling, PL performed best at $K = 4$, with a final mean success rate of $41\%$. PL averaged $22\%$ at $K = 3$ and $26\%$ at $K = 5$. BT-Kwise performed best at $K = 4$ with $45\%$, followed by $36\%$ at $K = 3$, but fell to $7\%$ at $K = 5$. Therefore, PL substantially outperformed BT-Kwise at $K = 5$, achieving $26\%$ compared with $7\%$. PL at $K = 4$ also exceeded the five-seed averages of RL-VLM-F at $28\%$ and BT-Pairwise at $38\%$. 

\subsubsection{Statistical Significance of Results}
To isolate the effect of the reward-learning objective, we compared PL and BT-Kwise at each matched ranking size using two-sided paired $t$-tests across the five corresponding seeds. No statistically significant difference was observed at any ranking size for Drawer Open ($K=3$, $p=0.436$; $K=4$, $p=0.883$; $K=5$, $p=0.607$) or Door Close ($K=3$, $p=0.836$; $K=4$, $p=0.839$; $K=5$, $p=0.830$). For Button Press, no significant difference was observed at $K=3$ ($p=0.307$) or $K=4$ ($p=0.845$). However, PL at $K=5$, achieved a higher mean final success rate than BT-Kwise, reaching $26\%$ compared with $7\%$, and this paired difference was nominally significant ($t(4)=2.881$, $p=0.045$) prior to a Holm correction ($p=0.405$ with correction). Thus, within the present five-seed evaluation, preserving the complete PL ranking likelihood did not yield a detectable final-success advantage for Drawer Open or Door Close, but it produced a nominally significant advantage over rank-breaking for Button Press at $K=5$ under the uncorrected test.

%decided against the overall table of k across all envinroments cause I think that defeats the point.

\subsection{Effects of Feedback Budget} 
% K=3,4,5 study (this hasn't been done) followed by the Drawer M=1 versus M=4 ablation.

\begin{figure}[h]
\centering
\includegraphics[width=\columnwidth]{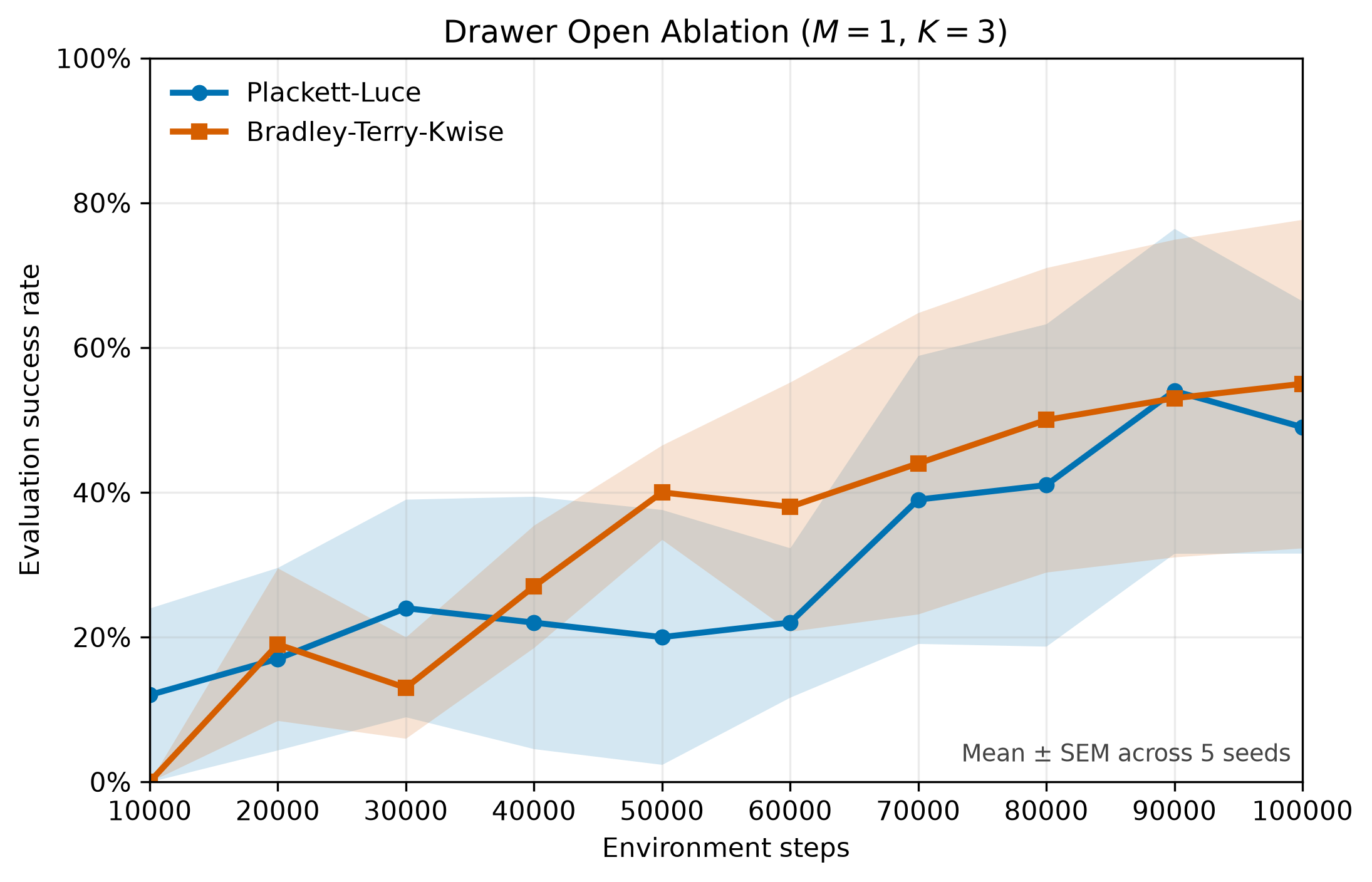}
\caption{Ablation aggregate success rate across five seeds $\pm$ standard error of the mean (SEM). Overall, Bradley-Terry had a higher average success rate after 100,000 environment steps, with a final success rate of 55\% compared to 49\% with Plackett-Luce.}
\label{fig:ablation}
\end{figure}

\begin{table}[h]
\centering
\caption{Ablation Final Success Rates per Run}
\resizebox{0.9\columnwidth}{!}{%
\begin{tabular}{|c|c|c|}
\hline
\textbf{Seed} & \textbf{PL Success-Rate} & \textbf{BT-Kwise Success-Rate} \\ \hline
0 & 0\% & 0\% \\ \hline
1 & 65\% & 95\% \\ \hline
2 & 15\% & 100\% \\ \hline
3 & 85\% & 0\% \\ \hline
4 & 80\% & 80\% \\ \hline
\textbf{Mean} & 49\% & 55\% \\ \hline
\end{tabular}%
}
\label{tab:my-table}
\end{table}

Having established that Plackett-Luce can learn an effective reward when four groups are sampled from the replay buffer per iteration ($M=4$), we next asked whether it could still learn a useful reward from only a single sampled group ($M=1$). On Drawer Open, reducing the feedback budget to $M=1$ resulted in final success rates of $0.49 \pm 0.17$ for PL and $0.55 \pm 0.23$ for BT-Kwise (mean $\pm$ SEM across five seeds). Performance was highly seed-dependent as both methods failed completely on at least one seed while attaining high or near-perfect success on others.

To contextualize these results, PL with $M=1$ achieved a lower mean success rate than the $M=4$ BT-Pairwise and RL-VLM-F baselines ($49\%$ versus $89\%$ and $92\%$, respectively). However, paired comparisons across five matched seeds did not meet the conventional threshold for statistical significance (BT-Pairwise versus PL: paired $t$-test, $p=0.093$; RL-VLM-F versus PL: $p=0.116$). Therefore, although the observed mean differences were large, the present experiment did not statistically establish a performance advantage for either $M=4$ method.

Moreover, PL with $M=1$ remained competitive with the highest-performing baseline on individual seeds. It outperformed RL-VLM-F on seed 3 ($85\%$ versus $80\%$) and matched it on seed 4 ($80\%$), thereby matching or exceeding RL-VLM-F on two of the five seeds. Notably, to achieve this, PL with $M=1$ required only 182 VLM calls, compared with 728 for BT-Pairwise and 1,456 for RL-VLM-F. VLM query details are listed in the appendix.

\section{Discussion}
\subsection{Key Findings and Implications}
% Interpret method, K, M, and query-efficiency results.
% Briefly discuss why results differ across tasks.
\subsubsection{Primary}

The results are also encouraging because PL produced successful policies under the relatively conservative $M=4$ feedback budget and 100,000 environment steps as compared to other VLM PbRL experiments \cite{wang2024rlvlmfreinforcementlearningvision}. This demonstrates that useful listwise reward learning is possible without requiring a large number of groups to compare at every iteration. The ability to vary $K$ gives practitioners direct control over the amount of information presented to the VLM to be jointly compared, thereby having control over the complexity of each ranking decision as well. For researchers and practitioners working on VLMs, this Plackett-Luce framework may offer a method to evaluate whether their model can, like humans, discern fine-grained differences among multiple candidate outputs based on the implications for the downstream policy outcome.

The matched comparison with BT-Kwise provides evidence that retaining the full listwise likelihood can be beneficial. The clearest result occurred for Button Press at $K=5$, where PL achieved a mean final success rate of $26\%$ compared with $7\%$ for BT-Kwise. This was the only matched comparison to produce a nominally significant difference across the five seeds under the uncorrected test, with $t(4)=2.881$ and $p=0.045$. Although this result should be interpreted cautiously because of the small sample size and multiple comparisons, it suggests that preserving the complete ranking structure can outperform rank-breaking when the ranking contains more candidates.

PL also compared competitively with the pairwise baselines across all environments. For Door Close, PL with $K=5$ achieved $54\%$ mean final success, compared with $39\%$ for both RL-VLM-F and single-stage BT-Pairwise. For Button Press, PL with $K=4$ achieved $41\%$, compared with $28\%$ for RL-VLM-F and $38\%$ for BT-Pairwise. RL-VLM-F primarily distinguishes itself from the single-stage pairwise baseline through its two-stage prompting procedure and additional VLM refinement calls while retaining uniform sampling \cite{wang2024rlvlmfreinforcementlearningvision}. The fact that PL can already match or exceed these pairwise methods without that additional prompting machinery suggests a natural direction for future work. A stronger system could combine PL's listwise objective with the multi-stage prompting and refinement strategy used by RL-VLM-F. This would preserve the efficiency and flexibility of listwise feedback while potentially improving the quality of the VLM-generated supervision.

\subsubsection{Feedback Group Ablation}

Our ablation results suggest that selecting multiple informative listwise groups may not always be necessary for effective reward learning. Prior active preference-learning methods search a large candidate buffer for informative comparisons using criteria such as reward-ensemble disagreement or preference entropy \cite{lee2021pebble}. Active subset selection has also been studied for Plackett-Luce models \cite{saha2019activerankingsubsetwisepreferences, thekumparampil2024comparingrankmanyactive}. In contrast, PL with $M=1$ obtains only a single uniformly sampled ranking of $K=3$ observations. Despite this restricted feedback budget, it matched or exceeded RL-VLM-F on two of five seeds, outperforming it on one seed and matching it on another, while using one-eighth as many VLM calls. This suggests that the breadth of a single $K$-wise comparison can sometimes provide sufficient supervision without extensive query
curation.

This result also motivates studying whether the warm-up period used to populate a large candidate buffer can be reduced. Since the ablation experiment does not search for multiple informative candidate groups, its results indicate maintaining a large buffer solely for query selection may be less critical. A shorter warm-up could reduce the amount of unguided exploratory behavior required before reward-guided policy training begins, which may make preference-based reward learning more practical and safer for real-world robotic systems. These findings motivate treating active acquisition as a potential enhancement rather than a strict prerequisite. Nevertheless, the lower mean performance and greater seed sensitivity observed under $M=1$ prevent it from serving as a robust replacement for higher-feedback configurations.

\subsection{Limitations}
% Few tasks/seeds, hosted-model drift, prompt/camera dependence, cost, time,
% simulator-only results, CNN capacity, and no human preferences.

First, identifying successful environments required multiple trials and various camera settings as the VLMs tested were sensitive to changes in the visual field and struggled with more complex scenarios, particularly grasping scenarios. This resulted in a relatively small number of evaluated tasks.
Second, preliminary experiments indicated that Qwen3-VL-8B-Instruct \cite{bai2025qwen3} did not yield any successful policies under our evaluation protocol, resulting in 0\% success in the tested Meta-World Reach task. As a result, reproducing our configuration requires inference costs associated with proprietary hosted VLMs. Lastly, we only simulated environments containing rigid-body tasks.

% Second, our early evaluation of PbRL in Meta-World environments indicated that open-source models were unable to provide meaningful preference labels compared to the proprietary GPT-5.6 Luna model. In particular, we tested Qwen3-VL-8B-Instruct \cite{bai2025qwen3} which produced a 0\% success rate over two test runs of PL and BT each on the Meta-World reach environment. As a result, reproducing our configuration requires inference costs associated with proprietary hosted VLMs. Furthermore, we only evaluate these methods in simulated environments which only contain rigid-body tasks.

\subsection{Future Work}
% More tasks, real robots, human labels, other VLMs, adaptive K/M,
% active query selection, and stronger visual reward encoders. mm also if we do even larger listwise selection, like does that increase the cost literally, and exploring the balance between that could be something
Future work can expand the diversity of tasks evaluated by exploring alternative benchmarks, soft-body simulations, and potentially dual-armed tasks. This could also include longer training runs beyond 100,000 environment steps and increase the number of feedback groups $M$. Furthermore, future work could explore larger listwise comparisons and examine the trade-off between sample efficiency and computational cost. Another interesting analysis would be to take the presented sampling methods to test on various reinforcement learning algorithms beyond SAC. Moreover, one could also explore how active selection of desirable pairs or sets of images to provide preference labels changes our presented results. To our knowledge, VLM-generated listwise PL reward learning has not yet been evaluated on a real-world robotic system. For this, it might be worth testing a reduced warm-up buffer and how that balances with active selection. 
% We plan on later evaluating VLM PbRL as a viable system in a surgical robotic context.

\section{Conclusions}

% In this paper, we present the first demonstration and validation of a listwise method for VLM preference-based reinforcement learning based on the Plackett-Luce reward model. Across three simulated robotic environments, the Plackett-Luce method was able to compete with or outperform both listwise (BT-Kwise) and pairwise (BT-Pairwise, RL-VLM-F) baseline methods. Results indicate that the Plackett-Luce method is a viable and competitive way of performing VLM preference-based reinforcement learning in object manipulation tasks. 

In this paper, to our knowledge, we present the first listwise VLM preference-based reinforcement learning framework using the Plackett-Luce reward model. Across three simulated robotic environments, PL successfully learned rewards from VLM generated $K$-wise rankings, achieving performance competitive with both BT-Kwise and the pairwise baselines of BT-Pairwise and RL-VLM-F. Results indicate that neither ranking size nor reward-model formulation provides a universal advantage, although preserving the full PL likelihood produced a nominally significant improvement over BT-Kwise for Button Press at $K=5$. Overall these findings establish listwise VLM supervision as a viable alternative to conventional pairwise preference feedback while providing flexibility over the amount and structure of supervision through both $K$ and $M$. Future work should explore adaptive ranking sizes, improved query selection, and evaluation on real-world robotic tasks.

\section*{AI Disclosure}
We used GPT-5.6, Claude Opus 5, and Claude Fable 5 to assist with code development and implementation. The tool materially affected algorithms in methods, experimental setup, and plot design in results. Reported data in results have been cross-checked and verified as correct.

\section*{Acknowledgments}
This material is based upon work supported by the National Science Foundation under Grant No. 2521982 and the NSF CISE REU Student Funding Program. We would also like to acknowledge Md Masudur Rahman for his assistance in reviewing this work.

\bibliographystyle{IEEEtran}
\bibliography{main}

\appendices
\newpage

\section{Query Efficiency and Pricing}
% Success per VLM call, total requests/tokens, abstentions,
% and VLM/oracle preference agreement.

\begin{figure}[h]
\centering
\includegraphics[width=0.9\columnwidth]{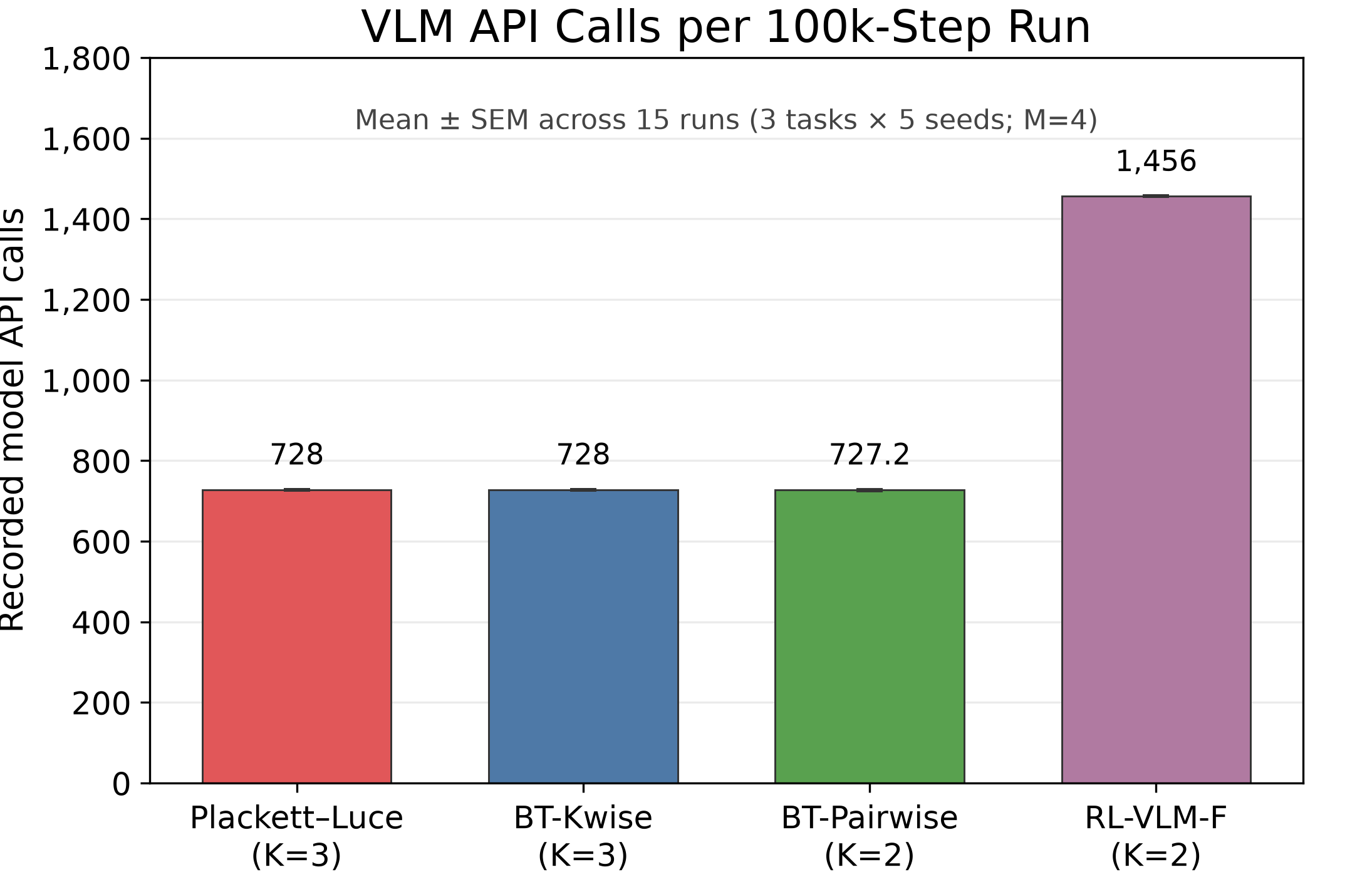}
\caption{Number of OpenAI API queries per run method}
\label{fig:api_calls}
\end{figure}

\begin{figure}[h]
\centering
\includegraphics[width=0.9\columnwidth]{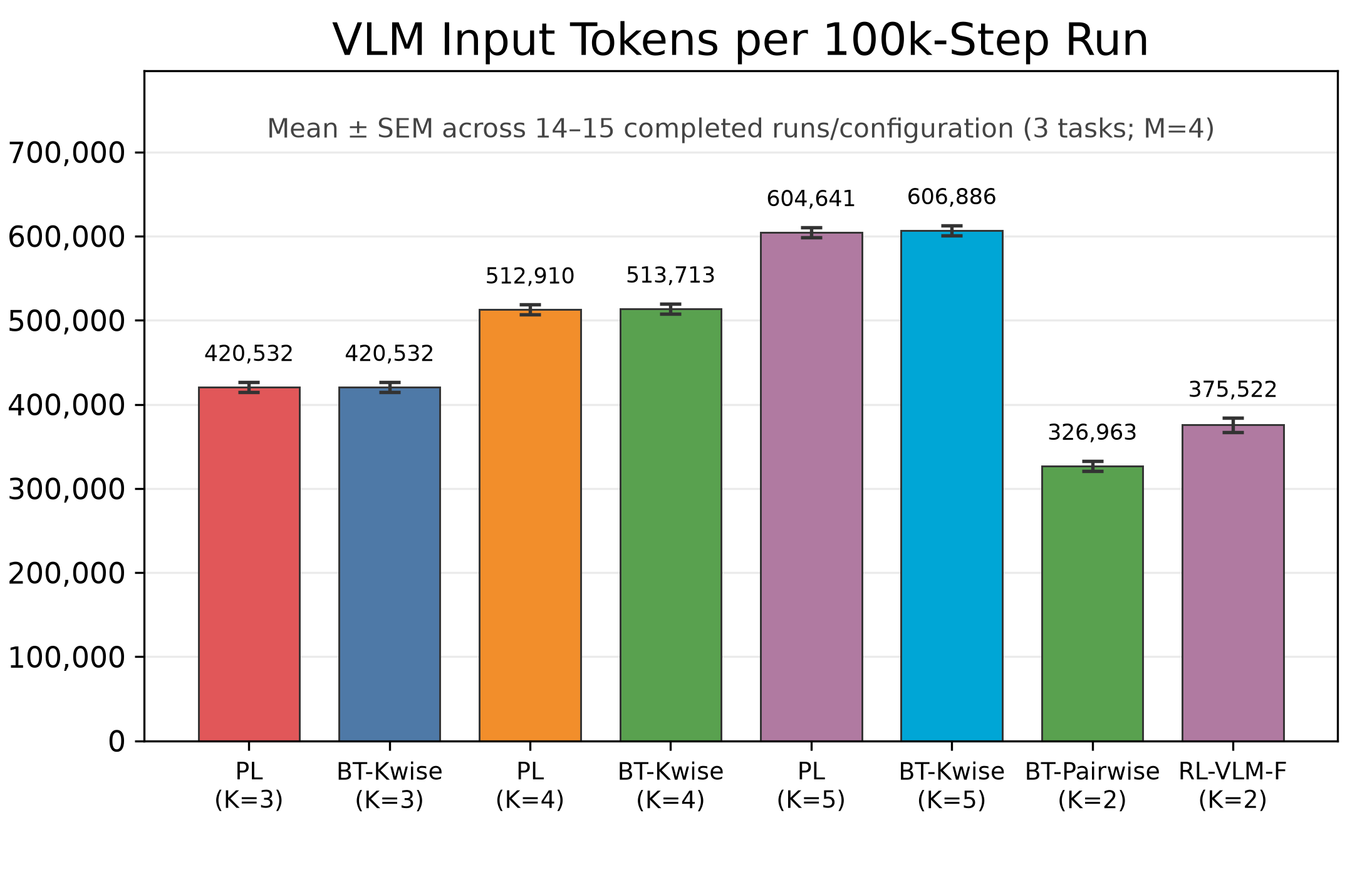}
\caption{Number of input tokens used per run method}
\label{fig:ablation}
\end{figure}

\begin{figure}[h]
\centering
\includegraphics[width=0.9\columnwidth]{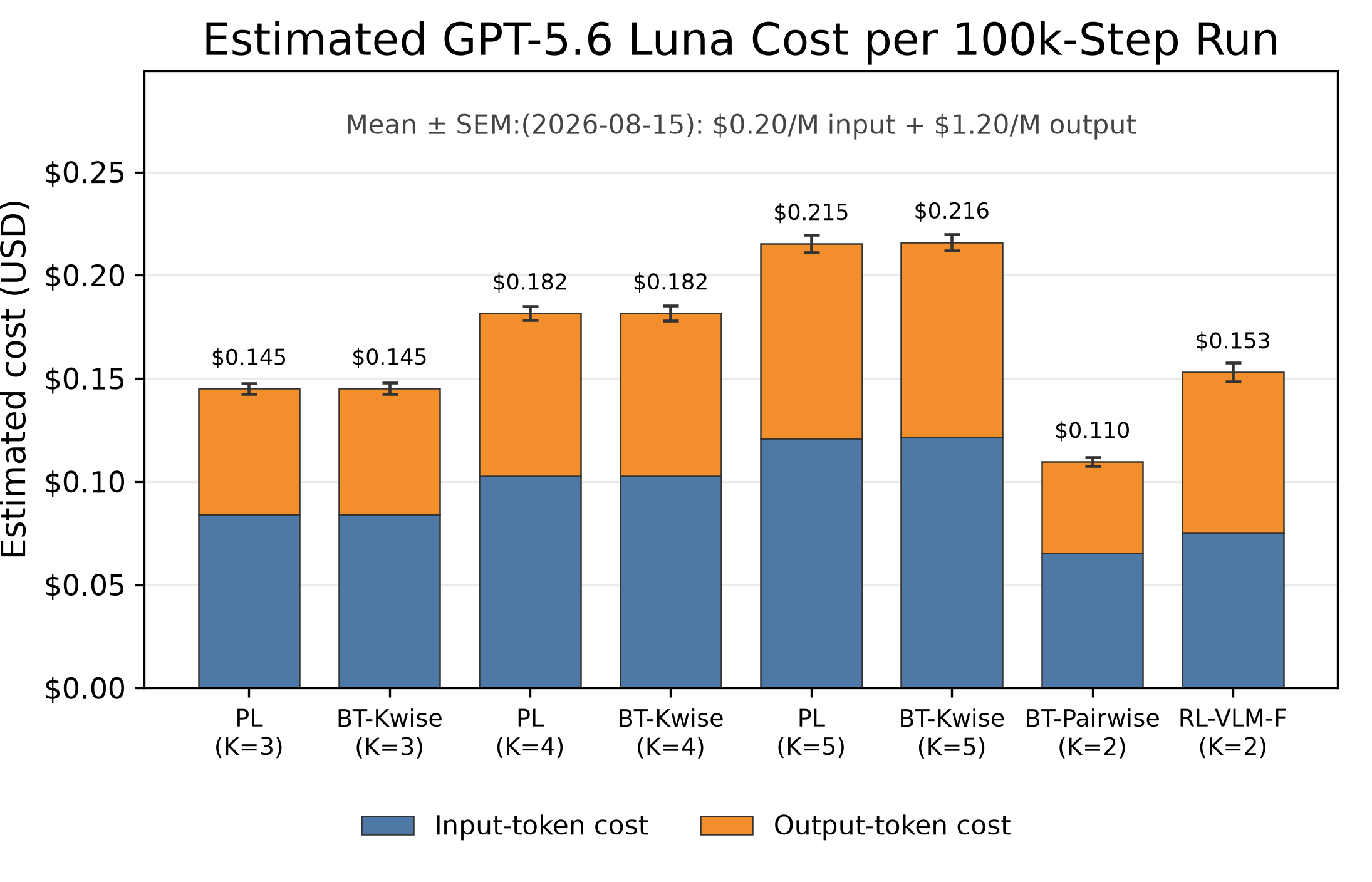}
\caption{Estimated cost of each run method with $M=4$. Estimations come directly from OpenAI's GPT-5.6 Luna pricing \cite{openai2026gpt56luna}.}
\label{fig:ablation}
\end{figure}

\section{Additional Results}

\subsection{Primary Results: Per Seed Success Rate}

\begin{table}[H]
\centering
\caption{Drawer Open Final Success Rates by Seed (\%).}
\label{tab:drawer_seed_results}
\resizebox{\columnwidth}{!}{%
\begin{tabular}{lcccccc}
\hline
\textbf{Method} &
\textbf{Seed 0} &
\textbf{Seed 1} &
\textbf{Seed 2} &
\textbf{Seed 3} &
\textbf{Seed 4} &
\textbf{Mean $\pm$ SEM} \\
\hline

PL ($K=3$)
& 100 & 75 & 80 & 35 & 80
& $74.0 \pm 10.7$ \\

PL ($K=4$)
& 85 & 85 & 80 & 80 & 100
& $86.0 \pm 3.7$ \\

PL ($K=5$)
& 85 & 75 & 45 & 100 & 100
& $81.0 \pm 10.2$ \\
\hline

BT-Kwise ($K=3$)
& 75 & 90 & 85 & 75 & 90
& $83.0 \pm 3.4$ \\

BT-Kwise ($K=4$)
& 85 & 85 & 100 & 100 & 50
& $84.0 \pm 9.1$ \\

BT-Kwise ($K=5$)
& 65 & 100 & 100 & 80 & 100
& $89.0 \pm 7.1$ \\
\hline

BT-Pairwise
& 100 & 90 & 75 & 80 & 100
& $89.0 \pm 5.1$ \\

RL-VLM-F
& 100 & 100 & 100 & 80 & 80
& $\mathbf{92.0 \pm 4.9}$ \\

Oracle
& 100 & 90 & 95 & 100 & 45
& $86.0 \pm 10.4$ \\
\hline

\end{tabular}%
}
\end{table}

\begin{table}[H]
\centering
\caption{Door Close Final Success Rates by Seed (\%).}
\label{tab:door_seed_results}
\resizebox{\columnwidth}{!}{%
\begin{tabular}{lcccccc}
\hline
\textbf{Method} &
\textbf{Seed 0} &
\textbf{Seed 1} &
\textbf{Seed 2} &
\textbf{Seed 3} &
\textbf{Seed 4} &
\textbf{Mean $\pm$ SEM} \\
\hline

PL ($K=3$)
& 90 & 5 & 40 & 60 & 5
& $40.0 \pm 16.4$ \\

PL ($K=4$)
& 50 & 10 & 100 & 80 & 0
& $48.0 \pm 19.3$ \\

PL ($K=5$)
& 60 & 35 & 15 & 70 & 90
& $\mathbf{54.0 \pm 13.2}$ \\
\hline

BT-Kwise ($K=3$)
& 60 & 55 & 0 & 100 & 5
& $44.0 \pm 18.7$ \\

BT-Kwise ($K=4$)
& 45 & 75 & 100 & 30 & 10
& $52.0 \pm 16.0$ \\

BT-Kwise ($K=5$)
& 10 & 25 & 100 & 100 & 0
& $47.0 \pm 22.0$ \\
\hline

BT-Pairwise
& 0 & 0 & 75 & 20 & 100
& $39.0 \pm 20.5$ \\

RL-VLM-F
& 45 & 40 & 0 & 15 & 95
& $39.0 \pm 16.2$ \\

Oracle
& 100 & 100 & 100 & 100 & 100
& $\mathbf{100.0 \pm 0.0}$ \\
\hline

\end{tabular}%
}
\end{table}

\begin{table}[H]
\centering
\caption{Button Press Final Success Rates by Seed (\%).}
\label{tab:button_seed_results}
\resizebox{\columnwidth}{!}{%
\begin{tabular}{lcccccc}
\hline
\textbf{Method} &
\textbf{Seed 0} &
\textbf{Seed 1} &
\textbf{Seed 2} &
\textbf{Seed 3} &
\textbf{Seed 4} &
\textbf{Mean $\pm$ SEM} \\
\hline

PL ($K=3$)
& 30 & 25 & 25 & 30 & 0
& $22.0 \pm 5.6$ \\

PL ($K=4$)
& 35 & 95 & 0 & 5 & 70
& $41.0 \pm 18.4$ \\

PL ($K=5$)
& 15 & 45 & 30 & 10 & 30
& $26.0 \pm 6.2$ \\
\hline

BT-Kwise ($K=3$)
& 40 & 85 & 20 & 25 & 10
& $36.0 \pm 13.2$ \\

BT-Kwise ($K=4$)
& 60 & 45 & 60 & 15 & 45
& $\mathbf{45.0 \pm 8.2}$ \\

BT-Kwise ($K=5$)
& 0 & 0 & 20 & 0 & 15
& $7.0 \pm 4.4$ \\
\hline

BT-Pairwise
& 0 & 70 & 50 & 0 & 70
& $38.0 \pm 15.9$ \\

RL-VLM-F
& 60 & 55 & 15 & 0 & 10
& $28.0 \pm 12.3$ \\

Oracle
& 60 & 100 & 100 & 100 & 80
& $\mathbf{88.0 \pm 8.0}$ \\
\hline

\end{tabular}%
}
\end{table}

\subsection{Effect of State-Diverse Sampling}

Our primary experiments use uniform sampling from the replay buffer. This is a deliberately conservative setting intended to test whether Plackett-Luce reward learning can succeed without active or diversity-aware query selection. Therefore, we aimed to obtain some initial insights into what may occur when using a form of active selection, state-diverse sampling, which builds ranked groups from observations that capture different stages of task progression.
\begin{table}[H]
\centering
\caption{Effect of State-Diverse Sampling on Drawer Open}
\label{tab:drawer_state_diverse}
\begin{tabular}{|c|c|c|}
\hline
\textbf{Seed}
& \textbf{Uniform}
& \textbf{State-Diverse} \\
\hline
0 & 0\%  & 80\%  \\
1 & 65\% & 100\% \\
2 & 15\% & 100\% \\
\hline
\textbf{Mean $\pm$ SEM}
& $\mathbf{27\% \pm 20\%}$
& $\mathbf{93\% \pm 7\%}$ \\
\hline
\end{tabular}
\end{table}
On Drawer Open, the potential benefit is stark under the restricted $M=1$, $K=3$ feedback budget. Across three matched seeds, uniform sampling achieves $27\% \pm 20\%$ success, whereas state-diverse sampling achieves $93\% \pm 7\%$. This suggests that selecting a behaviorally diverse ranking group can make a single VLM query substantially more informative.

\subsection{Effect of Increasing the Number of Feedback Groups}

Increasing $M$ provides the reward model with additional independent rankings at each feedback iteration. Door Close provides the clearest example of the benefit of increasing this feedback budget. Holding $K=3$ and state-diverse sampling fixed, increasing $M$ from one to five raises the mean success rate from $25\%$ to $62\%$ across the three matched seeds available in both settings. The number of rankings collected during training correspondingly increases from 182 to 910, providing substantially broader coverage of task progress and reducing the influence of any individual noisy VLM judgment.

\begin{table}[H]
\centering
\caption{Effect of Increasing Feedback Groups on Door Close}
\label{tab:door_close_increasing_m}
\begin{tabular}{|l|c|c|}
\hline
\textbf{Setting}
& \textbf{$M=1$}
& \textbf{$M=5$} \\
\hline
Ranking size, $K$       & 3   & 3   \\
Sampling strategy       & State-diverse & State-diverse \\
Total rankings          & 182 & 910 \\
\hline
Seed 0 success          & 65\% & 75\%  \\
Seed 1 success          & 0\%  & 100\% \\
Seed 2 success          & 10\% & 10\%  \\
\hline
\textbf{Mean $\pm$ SEM}
& $\mathbf{25\% \pm 20\%}$
& $\mathbf{62\% \pm 27\%}$ \\
\hline
\textbf{Mean improvement}
& \multicolumn{2}{c|}{$\mathbf{+37}$ percentage points} \\
\hline
\end{tabular}
\end{table}

This result suggests that increasing $M$ can improve performance, but at the cost of requiring more feedback queries. Our primary $M=4$ experiments are therefore conservative relative to methods that use much larger preference budgets. For example, the original RL-VLM-F formulation samples 20 pairs per iteration \cite{wang2024rlvlmfreinforcementlearningvision}, illustrating that strong pairwise reward-learning methods may rely on considerably more preference supervision. PL can similarly benefit from increasing $M$, while retaining the ability to order $K$ observations jointly within each ranking query.

\section{Alternative Run Configuration}

\begin{table}[H]
\centering
\caption{Alternate Configuration}
\resizebox{0.9\columnwidth}{!}{%
\begin{tabular}{|l|l|}
\hline
\textbf{OS} & Ubuntu 24.04.3 LTS \\ \hline
\textbf{GPU} & NVIDIA GeForce RTX 3090 (24 GB) \\ \hline
\textbf{NVIDIA Driver} & 580.173.02 \\ \hline
\textbf{NVCC Version} & V12.0.140 \\ \hline
\textbf{CPU} & 11th Gen Intel(R) Core(TM) i9-11900KF \\ \hline
\textbf{Python} & 3.12.13 \\ \hline 
\end{tabular}%
}
\label{tab:hardware_dvrk}
\end{table}

With this configuration, reward inference batch size was decreased from 512 to 128.

\section{CartPole Validation}

Prior to testing the Meta-World environments, we tested with OpenAI Gym's CartPole environment \cite{brockman2016openai} with an older version of our setup which utilized Claude Sonnet 5 \cite{ClaudeSonnet5}. Two test runs were performed with $K=4$ and $M=3$ on both the PL and BT-Kwise reward models. Similar to our presented configuration, the training begins with a warm-up phase for the first 7,500 environment steps and then performs VLM PbRL for a remaining 42,500 steps. Both tests resulted in a final success rate of 100\%, acting as an initial sanity test of the testing setup. 

\begin{figure}[h]
\centering
\includegraphics[width=0.9\columnwidth]{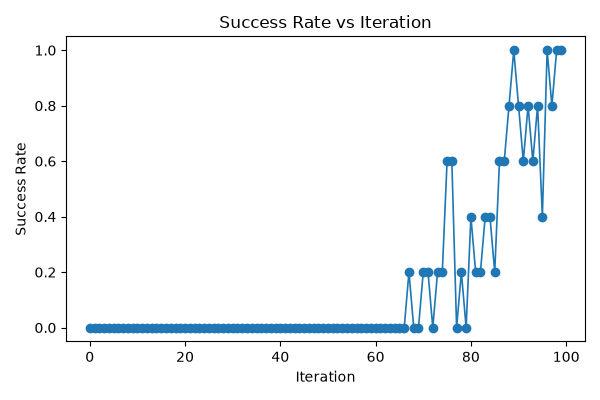}
\caption{Plackett-Luce CartPole success rate}
\label{fig:PL_CartPole}
\end{figure}

\begin{figure}[h]
\centering
\includegraphics[width=0.9\columnwidth]{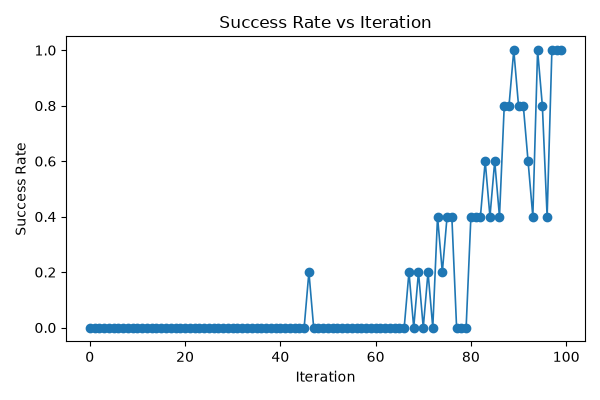}
\caption{Bradley-Terry-Kwise CartPole success rate}
\label{fig:BT_CartPole}
\end{figure}

\section{Full Hyperparameter List}

\label{app:hyperparameters}

Table~\ref{tab:hyperparameters} summarizes the hyperparameters used in the primary experiments. Unless otherwise specified, these settings were held constant across methods.

``Paper configuration'' refers to the following CNN design: RGB → Conv(16, 5×5, s=3) → Conv(32, 3×3, s=2) → Conv(64, 3×3, s=2) → Conv(128, 3×3, s=2) → Flatten → Linear(1) → tanh, with LeakyReLU after each convolution.

\begin{table}[h]
\centering
\caption{Primary Training and Evaluation Hyperparameters}
\label{tab:hyperparameters}

\scriptsize
\setlength{\tabcolsep}{3pt}
\renewcommand{\arraystretch}{0.95}

\begin{tabular}{@{}p{0.48\columnwidth}p{0.44\columnwidth}@{}}
\hline
\textbf{Parameter} & \textbf{Value} \\
\hline

\multicolumn{2}{l}{\textit{Vision-Language Model}} \\
VLM backend & OpenAI \\
VLM model & GPT-5.6 Luna \\
Image detail & Original \\
Reasoning effort & None \\
VLM timeout & 180 s \\
Query workers & 5 \\
Request limit & 80 requests/min \\
Candidate pool size & 1024 \\
Query sampling & Uniform \\
VLM label auditing & Enabled \\
\hline

\multicolumn{2}{l}{\textit{Reward Model}} \\
Reward input & Image \\
Vision encoder & CNN \\
CNN architecture & Paper configuration \\
Reward image size & $300 \times 300$ \\
Reward learning rate & $3\times10^{-4}$ \\
Reward ensemble size & 3 \\
Reward training epochs & 25 \\
Reward batch size & 40 \\
Early-stopping patience & 5 epochs \\
Preference validation fraction & 0.20 \\
Reward clipping & 5 \\
Reward inference batch size & 512 \\
Precompute iteration rewards & Enabled \\
Reward-frame caching & Disabled \\
\hline

\multicolumn{2}{l}{\textit{Warm-up and Replay Buffer}} \\
Warm-up iterations & 18 \\
Warm-up objective & State entropy \\
Ground-truth training reward & Disabled \\
Critic reset after warm-up & Enabled \\
Replay-buffer capacity & 100,000 \\
Frame storage & JPEG \\
JPEG quality & 95 \\
\hline

\multicolumn{2}{l}{\textit{Soft Actor-Critic}} \\
SAC batch size & 512 \\
Updates per environment step & 1.0 \\
Policy learning rate & $3\times10^{-4}$ \\
Critic learning rate & $3\times10^{-4}$ \\
Temperature learning rate & $1\times10^{-4}$ \\
Initial temperature ($\alpha$) & 0.1 \\
Critic loss & Huber \\
Target Q clipping & 200 \\
Gradient clipping norm & 10 \\
Maximum critic loss & 1000 \\
Policy hidden architecture & 3 layers, 256 units each \\
Critic hidden architecture & 3 layers, 256 units each \\
Target-update frequency & 2 \\
\hline

\multicolumn{2}{l}{\textit{Environment and Training}} \\
Environment & Meta-World \\
Tasks & Drawer Open, Door Close, Button Press \\
Maximum episode length & 500 steps \\
Training iterations & 200 \\
Rollouts per iteration & 1 \\
Total environment steps & 100,000 \\
Random seeds & 0--4 \\
Render resolution & $300 \times 300$ \\
Render rotation & $180^\circ$ \\
\hline

\multicolumn{2}{l}{\textit{Evaluation and Logging}} \\
Evaluation interval & 20 iterations (10,000 steps) \\
Evaluation episodes & 20 \\
Checkpoint interval & 20 iterations \\
Video interval & 20 iterations \\
Video frame stride & 5 \\
\hline

\end{tabular}
\end{table}

\section{Task Prompts}
The below task prompts are injected into the overall goal prompt in the next section for PL, BT-Kwise, and BT-Pairwise.
\subsection{Drawer Open}
Open the drawer by pulling it outward; prefer the image in which the drawer is visibly farther open.

\subsection{Door Close}
Close the dark hinged door against its cabinet frame. Prefer the image with a smaller door opening and the door more nearly flush with the cabinet. Judge the door angle, not gripper position.

\subsection{Button Press}
Press the red circular button fully into the yellow box. Rank primarily by the length of the exposed dark gray shaft between the red button cap and the yellow box: a shorter exposed shaft means the button is farther pressed and is always better; a red cap flush against the yellow face is best. Ignore fingertip position whenever shaft lengths differ. Only when shaft lengths are visually tied, prefer the white-and-cyan fingertips closer to, aligned with, and pushing the red cap.

\section{Goal Prompt}
The goal is to [task description]. Rank the images from best to worst by how well they achieve the goal. You must choose a complete best-to-worst ranking even if differences are subtle.

\section{RL-VLM-F Prompt}
RL-VLM-F uses the same task descriptions but has its own prompts for its two stage process.
\subsection{Visual Analysis}
Consider the following two robot end-effector observations.

[Image 1]

[Image 2]

\begin{enumerate}
    \item Describe what is happening in Image 1.
    \item Describe what is happening in Image 2.
    \item The goal is to \{TASK\_DESCRIPTION\}. Is there a difference between them in terms of how well the goal is being achieved?
\end{enumerate}

\subsection{Preference Selection}
Based on the text below answering the questions:

\{STAGE\_1\_RESPONSE\}

Is the goal better achieved by Image 1 or Image 2, all things considered?
Reply with a single line containing only 0 if Image 1 is better, or 1 if Image 2 is better. Reply -1 if you are unsure or there is no meaningful difference.

\end{document}